\documentclass[letterpaper, 10 pt, conference]{ieeeconf}
\IEEEoverridecommandlockouts
\usepackage{amsmath,amssymb,amsfonts}
\usepackage{algorithmic}
\usepackage{graphicx}
\usepackage{textcomp}
\usepackage{xcolor}
\usepackage{algorithm}
\usepackage{array}
\usepackage[caption=false,font=normalsize,labelfont=sf,textfont=sf]{subfig}
\usepackage{stfloats}
\usepackage{url}
\usepackage{verbatim}
\usepackage{times}
\usepackage{epsfig}
\usepackage{float}
\usepackage{wrapfig}
\usepackage{bm,xspace}
\usepackage{comment}
\usepackage{multirow}
\usepackage{balance}
\usepackage{booktabs}
\usepackage{etoolbox,siunitx}
\usepackage{calc}
\usepackage{pifont,hologo}
\usepackage{nicefrac}
\usepackage{makecell}
\usepackage{svg}
\usepackage{enumerate}

\makeatletter
\let\NAT@parse\undefined
\makeatother
\usepackage[colorlinks=black,citecolor=green]{hyperref}
\usepackage{adjustbox}

\def\BibTeX{{\rm B\kern-.05em{\sc i\kern-.025em b}\kern-.08em
    T\kern-.1667em\lower.7ex\hbox{E}\kern-.125emX}}

\title{\LARGE \bf Semi-distributed Cross-modal Air-Ground  Relative Localization}

\author{Weining Lu$^{1}$, Deer Bin$^{2}$, Lian Ma$^{2}$, Ming Ma$^{2}$, Zhihao Ma$^{2}$, Xiangyang Chen$^{3}$,\\ Longfei Wang$^{3}$, Yixiao Feng$^{1}$, Zhouxian Jiang$^{2}$, Yongliang Shi$^{2*}$, Bin Liang$^{2*}$
\thanks{$^{1}$Beijng National Research Center for Information Science and Technology, $^{2}$Qiyuan Lab, $^{3}$JiangHuai Advanced Technology Center.}%
\thanks{$*$ Corresponding author.}%
\thanks{Sponsored by the BNRist project (No. BNR2024TD03003) and Qiyuan Innovation Fund (NO.2022-JCJQ-LA-001-107).}
}

\begin{document}
\maketitle
\thispagestyle{empty}
\pagestyle{empty}

\begin{abstract}
Efficient, accurate, and flexible relative localization is crucial in air-ground collaborative tasks. However, current approaches for robot relative localization are primarily realized in the form of distributed multi-robot SLAM systems with the same sensor configuration, which are tightly coupled with the state estimation of all robots, limiting both flexibility and accuracy. To this end, we fully leverage the high capacity of Unmanned Ground Vehicle (UGV) to integrate multiple sensors, enabling a semi-distributed cross-modal air-ground relative localization framework. In this work, both the UGV and the Unmanned Aerial Vehicle (UAV) independently perform SLAM while extracting deep learning-based keypoints and global descriptors, which decouples the relative localization from the state estimation of all agents. The UGV employs a local Bundle Adjustment (BA) with LiDAR, camera, and an IMU to rapidly obtain accurate relative pose estimates. The BA process adopts sparse keypoint optimization and is divided into two stages: First, optimizing camera poses interpolated from LiDAR-Inertial Odometry (LIO), followed by estimating the relative camera poses between the UGV and UAV. Additionally, we implement an incremental loop closure detection algorithm using deep learning-based descriptors to maintain and retrieve keyframes efficiently. Experimental results demonstrate that our method achieves outstanding performance in both accuracy and efficiency. Unlike traditional multi-robot SLAM approaches that transmit images or point clouds, our method only transmits keypoint pixels and their descriptors, effectively constraining the communication bandwidth under 0.3 Mbps. Codes and data will be publicly available on \href{https://github.com/Ascbpiac/cross-model-relative-localization.git}{https://github.com/Ascbpiac/cross-model-relative-localization.git}.
\end{abstract}

\section{INTRODUCTION}
Relative localization is a fundamental problem in heterogeneous multi-robot systems which provides benefits in a wide range of tasks \cite{rizk2019cooperative, LINDQVIST2022104134,SPOMP}.
Current SLAM-based relative localization methods \cite{xu2024d, xu2022omni,zhu2023swarm,dube2017online} require significant data transmission, making them challenging to work effectively in bandwidth-limited environments. Additionally, their tight coupling with SLAM significantly limits the flexibility of relative localization methods, making them unsuitable for multi-robot systems with heterogeneous sensor configurations, and thereby hindering the execution of downstream tasks. 

Leveraging the UAV's extensive field of view and the UGV's superior payload capacity to support additional sensors and computational platforms, our relative localization framework is designed to be decoupled from individual SLAM systems in the multi-robot setting and can be flexibly integrated as a plugin into air-ground collaborative tasks with different SLAM approaches. 

For \textbf{efficiency} and \textbf{flexibility} of our work, UAV transmits sparse keypoints and corresponding descriptors \cite{xfeat} of its current observation image, along with the global descriptor \cite{mixvpr} for place recognition and camera pose (obtained via VIO \cite{geneva2020openvins}), to UGV. Similarly, UGV performs the same process on its observation image. However, its camera pose is obtained through linear interpolation of LIO states \cite{chen2023direct}. Notably, since the map initialization process of the selected LIO, similar to SLAM on UAV, takes the gravity direction into account, the z-axis of the map is always parallel to the gravity direction. Consequently, when estimating the rotational component of the relative pose between aerial and ground systems, only the yaw angle optimization needs to be considered. This reduces the dimensionality of state estimation and improves the efficiency of the estimation process.

For \textbf{accuracy}, a hash table \cite{fasterlio} of the dense local map is maintained to retrieve the depth of image keypoints from UGV. Once establishing correspondences between keypoints descriptor \cite{xfeat} of UGV and UAV \cite{douze2024faiss}, the relative pose is estimated using EPnP \cite{epnp}. It is evident that this approach alone is prone to significant errors, thus a two-stage optimization is introduced to correct the errors. In the first stage, bundle adjustment of the UGV's local sliding window is employed to correct camera pose errors caused by multi-sensor extrinsic and time offsets, as well as LiDAR depth information corresponding to keypoint pixels. In the second stage, air-ground visual-geometric consistency and relative pose constraint regularization are utilized to optimize the relative pose and camera poses of the UAV.

Furthermore, in the work of the air-ground system, the absence of a common field of view is inevitable. To facilitate long-term relative pose estimation, a database maintenance system of deep learning-based keyframe descriptors is developed, where an incremental Hierarchical Navigable Small Worlds (HNSW) \cite{nsw,hnsw} is employed to manage the keyframes on the UGV. These keyframes, represented as global descriptors \cite{mixvpr}, are derived from both the UAV and UGV keyframes, allowing for the retrieval of co-visibility information from historical frames for the subsequent stage of relative pose estimation.

In summary, our contributions are as follows:
\begin{itemize}
\item[$\bullet$] We propose a semi-distributed cross-modal relative localization architecture for air-ground systems, which decouples the state estimation of individual agents from the overall system. 
\item[$\bullet$] A two-stage optimization formulation is proposed to achieve efficient and accurate relative state estimation.
\item[$\bullet$] A database maintenance and retrieval system is proposed to build an incrementally updated database utilizing learning-based global descriptors for place recognition.
\item[$\bullet$] Datasets are published proving the superiority in efficiency, accuracy, and flexibility of our method over existing approaches.
\end{itemize}

\section{Related Work}
\subsection{Place Recognition}
The standard pipeline for localization approaches begins with retrieving prior information, termed place recognition. 
The robustness, interference resistance, and accuracy of the CNN-based image retrieval algorithm have been substantiated \cite{netvlad}. 
To enable deployment on mobile devices, a more lightweight visual place recognition approach is proposed \cite{mixvpr}. Once the reference images are retrieved, local feature extraction and matching are performed, then the Perspective-n-Point (PnP) algorithm \cite{kukelova2013real} combined with RANSAC is leveraged to estimate the 6-DoF pose. Nonetheless, single-modality approaches exhibit limitations when confronted with challenging scenarios, such as cross-seasonal variations and nighttime conditions.
i3dLoc \cite{yin2021i3dloc} presents an end-to-end visual place retrieval method with respect to offline 3D maps in varying environmental conditions. It retains condition-invariant features by a Generative Adversarial Network (GAN).
Tuzcuoğlu et al. \cite{2024xoftr} introduce a cross-modal, cross-view feature matching method between thermal infrared (TIR) and visible images facilitating place retrieval capabilities in nighttime scenarios.
Additionally, leveraging higher-level semantic information beyond geometric and textural features enhances both the robustness and discriminative capability of place recognition.
Miller utilizes semantics to globally localize a robot using egocentric 3D semantically labelled LiDAR and IMU as well as top-down RGB images obtained from satellites or aerial robots \cite{9361130}.
Gao \cite{gao2023visual} proposes a two-stage place recognition approach integrating a graph representation of spatial relationship of semantics and class-wise distance fields which encode geometrical cues. 
LIP-Loc \cite{shubodh2024lip} transfers Contrastive Language-Image Pre-Training (CLIP) to the domains of 2D image and 3D LiDAR points to achieve the multi-modal place recognition.
However, unlike \cite{galvez2012bags}, these approaches are devoid of a comprehensive and systematic framework for database management and retrieval.

\subsection{Cross-modal Relative Localization}
Accurate ego-state and relative state estimation are crucial prerequisites for multi-robot missions. Distributed multi-robot SLAM systems with homogeneous sensor setups have been proposed \cite{dube2017online, xu2022omni, zhu2023swarm,xunt2023crepes, xu2024d}.
Dubé et al. \cite{dube2017online} construct a centralized Pose Graph Optimization (PGO) utilizing sequential and place recognition constraints for a multiple cooperative robot system. 
CREPES \cite{xunt2023crepes} also leverages a PGO module to integrate the measurements from IR LEDs, cameras, UWBs, and IMUs, achieving highly accurate and robust relative localization, while it imposes significant requirements for hardware customization. 
\cite{xu2022omni, zhu2023swarm, xu2024d} develop decentralized collaborative SLAM systems with different sensor configurations (visual-inertial \cite{xu2024d}, lidar-inertial \cite{zhu2023swarm}, omnidirectional visual-inertial-UWB \cite{xu2022omni}). Despite these advances, the inability to support heterogeneous sensor setups limits system flexibility. 

Güler \cite{GULER2023105492} utilizes a leader-follower formation where the aerial robot is equipped with UWB and a camera, while ground robots use only UWB.
In SPOMP \cite{SPOMP}, independent odometry is realized through RGB cameras on the aerial agent and LiDAR on the ground agent, complemented by intermittent semantic information exchange for relative localization. However, they typically integrate relative localization among agents as a critical constraint for their state estimation process. 

In contrast, our work is deliberately decoupled from the individual odometry techniques employed by each agent, which enhances the flexibility and interoperability of the system, allowing agents utilizing diverse SLAM algorithms to effectively perform downstream collaborative tasks.

\section{Method}

\subsection{System Overview}
\begin{figure*}[ht]
\centering
\includegraphics[width=0.9\linewidth]{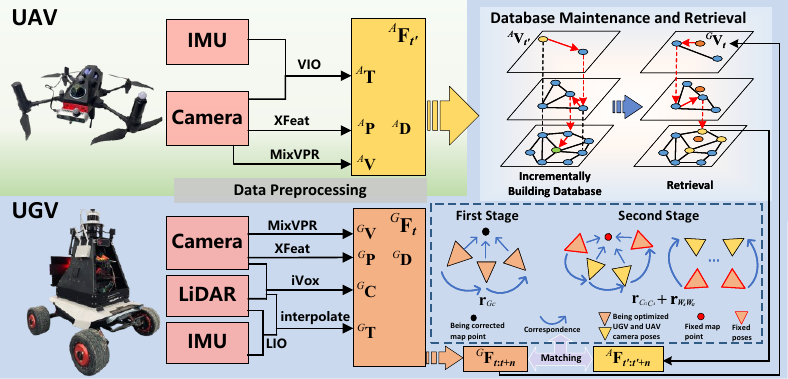}
\caption{The system consists of three parts: Data Preprocessing, Database Maintenance and Retrieval, and Two-stage Optimization.}
\label{fig:main}
\end{figure*}
As shown in Fig.\ref{fig:main}, we aim to estimate the state \( \mathcal{S} = \{\textbf{T}_{W_GW_A}, \{\textbf{T}_{W_AC_A}\}, \{\textbf{T}_{W_GC_G}\}\} \) and \( \mathcal{Z} = \{^{ij}\mathbf{C}\} \), where \( \textbf{T}_{W_GW_A} \) refers to the relative pose in the world coordinate system between the air and ground frames, \( \{\textbf{T}_{W_AC_A}\} \) represents the camera poses in the world coordinate system as measured by the UAV, while  \( \{\textbf{T}_{W_GC_G}\} \) denotes the same for the UGV, and \( \{^{ij}\mathbf{C}\} \) denotes the depth values corresponding to the $j$-th visual feature point in frame $i$ at the UGV end.
 UAV sends frame $^A\mathbf{F}$, 
 including sparse keypoint pixels $^{A}\mathbf{P}=\{^{A}\mathbf{p}_1,^{A}\mathbf{p}_2,\ldots,^{A}\mathbf{p}_n\}$, corresponding descriptors $^{A}\mathbf{D}=\{^{A}\mathbf{d}_1,^{A}\mathbf{d}_2,\ldots,^{A}\mathbf{d}_n\}$, global descriptor $^{A}\boldsymbol{V}$ for place recognition and camera pose $^{A}\boldsymbol{T}$. And the UGV will maintain a local sliding window of frame $^G\mathbf{F}=\{\{^{G}\mathbf{P}\}, \{^{G}\mathbf{D}\}, ^{G}\boldsymbol{T}, ^{G}\boldsymbol{V},\{^{G}\mathbf{C}\}\}$, of which image is processed in the same manner to obtain keypoints and descriptors $\{^{G}\mathbf{P}, ^{G}\mathbf{D},^{G}\boldsymbol{V}\}$, except that the camera pose $^{G}\boldsymbol{T}$ is obtained through linear interpolation of the LIO poses. Furthermore, the points $\{\mathbf{C}\}$ corresponding to the pixels are retrieved from the local point cloud map (Section \ref{sec:corr}). On the UGV side, while maintaining its own observation sliding window, it incrementally builds the database with $^{A}\boldsymbol{V}$s acquired from the UAV, facilitating the retrieval of co-visibility information from the UAV's historical observations $^{A}\boldsymbol{V}$s (Section \ref{sec:database}). After obtaining the most similar frames $^A\mathbf{F}$s of UAV, the UGV first uses its own visual spatiotemporal consistency of $^G\mathbf{F}$s to correct the camera poses $\{\textbf{T}_{W_GC_G}\}$ and spatial location of map points $\{\mathbf{C}\}$ (Section \ref{firstage}). Subsequently, optimization of camera poses $\{\textbf{T}_{W_GC_A}\}$ on UAV is performed through data association between air and ground-based cameras. To ensure numerical stability, the relative pose consistency of different air-ground image pairs is also regarded as a constraint to optimize the relative localization $\textbf{T}_{W_GW_A}$ (Section \ref{secstage}).
 

\subsection{Pixel-Point Correspondence}
\label{sec:corr}
Due to the sparsity of the point cloud of a single LiDAR frame, the keypoint pixels on the UGV index the corresponding depth values from the dense map of LiDAR-SLAM. 
\begin{figure}[h]
\centering
\includegraphics[width=0.75\linewidth]{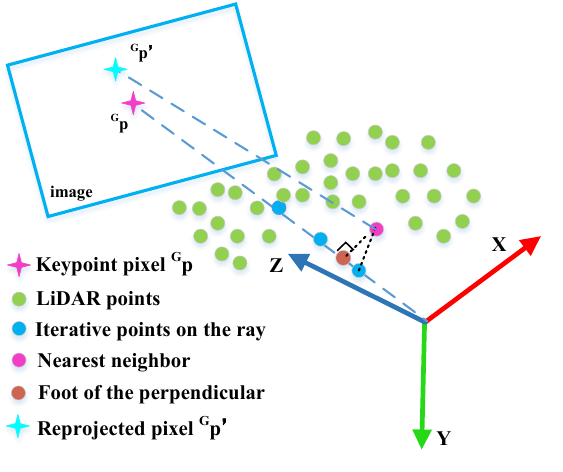}
\caption{Pixel-Point Correspondence.}
\label{fig:corr}
\vspace{-0.1in}
\end{figure}


Compared to the complete voxelization of octree voxel structures in ray casting \cite{fastlivo}, iVox \cite{fasterlio} offers a sparse voxel representation, establishing and maintaining voxel structures only where spatial points exist. Given the positions of spatial points, a hash table can be employed to uniformly construct and maintain the voxel structure, using the coordinates of the spatial points as keys and the unique indices generated by feeding these coordinates into a hash function as values. As the Fig.\ref{fig:corr} depicted, given a pixel \(^{G}\mathbf{p}\), a ray generated in camera coordinates is iterated along a specified depth range with a fixed step of 0.05 m, starting from shallower depths. At each step, the corresponding points are transformed to world coordinate. Concurrently, in a separate thread, point clouds are accumulated in a hash table, with world coordinates quantized based on a voxel size of 0.1 m. For each converted point, the nearest neighbor is retrieved from the hash table. If no neighbor is found, the iteration proceeds to deeper points. When a neighbor is identified, the reprojected pixel \(^{G}\mathbf{p'}\) is calculated, the reprojection error between \(^{G}\mathbf{p}\) and \(^{G}\mathbf{p'}\) is evaluated. If the error exceeds a threshold, the iteration continues; otherwise, the nearest neighbor is transformed back to camera coordinates, and the perpendicular foot from the point  to the ray serves as the precise 3D representation of the image feature pixel in camera coordinates.

\subsection{Database Maintenance and Retrieval}
\label{sec:database}
Although DBoW \cite{galvez2012bags} is widely utilized for incremental place recognition and retrieval in visual SLAM, its reliance on hand-crafted descriptors, such as SURF \cite{surf} and ORB \cite{orb}, renders it incompatible with deep learning-based descriptors, making it unsuitable for the requirements of this study. Given the demonstrated efficiency and accuracy of MixVPR \cite{mixvpr} in visual place recognition, coupled with the capability of HNSW \cite{hnsw} to support incremental graph construction, we adopt these methodologies to facilitate keyframe data management and retrieval. 

We first quantize the MixVPR into an FP16-format engine, which not only reduces the model's memory but also significantly enhances inference speed. Given the $^A\boldsymbol{V}$ by the MixVPR from the UAV, the HNSW index is constructed incrementally by inserting it into a multi-layered graph. 
Construction begins at the topmost layer, which has the fewest nodes. For each new $^A\boldsymbol{V}$, the algorithm performs a greedy search from an entry point to identify its nearest neighbors in the current layer. This process iterates downward through all layers up to the element’s maximum layer.

In the retrieval process, HNSW employs a multi-layered traversal to navigate the $^G\boldsymbol{V}$ to the most similar $^A\boldsymbol{V}$. The search initiates at the top layer, using a best-first strategy to navigate toward the query’s approximate neighbors. 
This hierarchical approach reduces the search complexity to approximately logarithmic scale relative to dataset size, as long-range links in higher layers prevent exhaustive traversal. 

\subsection{Two-stage Optimization}
\label{2stage}
Air-ground collaborative tasks require execution within a unified global coordinate system. Therefore, the objective of relative localization in this work is to align all agents within the global coordinate system, specifically to determine \( T_{W_AW_G} \) (eq. \eqref{rloc}).
\begin{equation}
\label{rloc}
T_{W_AW_G}=T_{W_AC_A}T_{C_AC_G}T^{-1}_{W_GC_G}
\end{equation}
Here, \( T_{W_AC_A} \) represents the pose of the UAV-mounted camera in the UAV's world coordinate system, \( T_{W_GC_G} \) denotes the pose of the UGV-mounted camera in the UGV's world coordinate system, and \( T_{C_AC_G} \) indicates the relative pose of the UGV-mounted camera with respect to the UAV-mounted camera. 
The transformation \( T_{W_GC_G} \) is derived via linear interpolation of the LiDAR odometry of the UGV. However, due to the slight variations in the extrinsic parameters between the LiDAR and camera during the movement of the UGV, as well as the lack of precise temporal synchronization between the sensors, certain errors exist in the camera poses obtained through linear interpolation of the LiDAR poses. Additionally, the depths associated with image keypoint pixels are obtained from the map of LiDAR SLAM, and leveraging these points to solve the Perspective-n-Point (PnP) inherently contributes to errors in the estimation of \(T_{W_AW_G}\). 
Additionally, the vio errors inherent to the UAV \(T_{W_AC_A}\) can introduce system inaccuracies, leading to instability in \(T_{W_AW_G}\). To address the aforementioned issue, we implemented a two-stage optimization process. Specifically, given the key points of the UGV and UAV, along with their corresponding depths on the UGV, we first optimize the camera poses and depths on the UGV. Subsequently, we optimize the relative pose $\mathbf{T}_{W_AW_G}$ and the camera poses $\{T_{W_AC_A}\}$ on the UAV simultaneously.

\subsubsection{\textbf{First Stage}}
\label{firstage}
Within a fixed-length sliding window, visual keyframes $^G\mathbf{F}_{t:t+n}$, assisted by lightweight and robust feature descriptors $\{^G\mathbf{D}\} $\cite{xfeat}. Therefore, local Bundle Adjustment (BA) \cite{pixelsfm,colmap} is introduced for pose optimization of the keyframes and correction of map point positions. As a gold standard, BA is for the refining structure and initial poses, leveraging the dense correspondences by the \cite{douze2024faiss}, it  minimizes the reprojection errors: 
\begin{equation}
 \label{ugvreproj}
     \mathbf{r}_{G_C}=\sum_{j} \sum_{(i, u) \in \mathcal{T}(j)}\left\|^GK\left(^j\mathbf{T}^{-1}_{W_GC_G} \mathbf{C}_i\right)-^G\mathbf{p}_u\right\|_\gamma
 \end{equation}
where $\mathcal{T}(j)$ is the set of frames $^G\mathbf{F}$ and keypoints corresponding the local map point index $i$, $\| \cdot \|_\gamma$ is a robust norm, $^GK$ is intrinsic parameters of the camera on UGV, then more accurate $\{\mathbf{T}_{W_GC_G}\}$ and $\{\mathbf{C}_i\}$ will be refined. Specially, to ensure numerical stability during the optimization process, we optimize the 3D map points using inverse depth parametrization.
In doing so, we implicitly correct the errors introduced by extrinsic parameters, time unsynchronization, and linear interpolation. 

\subsubsection{\textbf{Second Stage}}
\label{secstage}
Given the refined $\{\mathbf{T}_{W_GC_G}\}$ and $\{\mathbf{C}_j\}$, they are fixed in the subsequent optimization process. This stage establishes a nonlinear optimization incorporating dual constraint mechanisms, with the objective function formulated as:
\begin{equation}
\mathbf{r}=\mathbf{r}_{C_GC_A}+\mathbf{r}_{W_AW_G}
\end{equation}
$\mathbf{r}_{C_GC_A}$ denotes the reprojection error term  (eq. \eqref{agreproj}) derived from air-ground cross-platform dense feature correspondences, which simultaneously integrates spatial geometric constraints and multi-view texture consistency metrics. To get the correspondences, with the method from section \ref{sec:database}, the $\{^G\mathbf{F}_{t:t+n}\}$ on the UGV retrieve the most similar UAV frames $\{^A\mathbf{F}_{t':t'+n}\}$ from the database, and further filters reliable frames by using FAISS \cite{douze2024faiss} for matching, discarding pairs with fewer matches than a predefined threshold. 
\begin{equation}
\label{agreproj}
\begin{split}
\mathbf{r}_{C_GC_A} = \sum_{j} \sum_{(i, u) \in \mathcal{T}(j)} \Big\| {}^A K \big(&{}^j \mathbf{T}^{-1}_{W_AC_A}\mathbf{T}_{W_AW_G}  \\
&\mathbf{T}_{W_GC_G} \mathbf{C}_i \big) - {}^A \mathbf{p}_u \Big\|_\gamma
\end{split}
\end{equation}
It means that the \(\{\mathbf{T}_{W_AW_G}\}\) and camera poses of the UAV are both adjusted within the sliding window. Additionally, to enhance the robustness of pose graph optimization, we construct the relative pose constraint regularization term $\mathbf{r}_{W_GW_A}$ (eq. \eqref{gcon}), which enforces compatibility conditions on the Lie manifold for relative poses \(\{\mathbf{T}_{W_AW_G}\}\) derived from all air-ground keyframe pairs. By minimizing the residual $\mathbf{r}$, we simultaneously adjust \(\{\mathbf{T}_{W_AC_A}\}\) to optimize and obtain an accurate $\mathbf{T}_{W_AW_G}$.
\begin{equation}
\label{gcon}
\mathbf{r}_{W_AW_G}=\sum_{j}{\|\mathbf{T}_{W_AW_G}{^{j}\hat{\mathbf{T}}^{-1}_{W_AW_G}}-\mathbf{I}\|}
\end{equation}

\section{EXPERIMENTS AND ANALYSIS}
\subsection{Platform}
\begin{figure}[ht]
\centering
\includegraphics[width=0.85\linewidth]{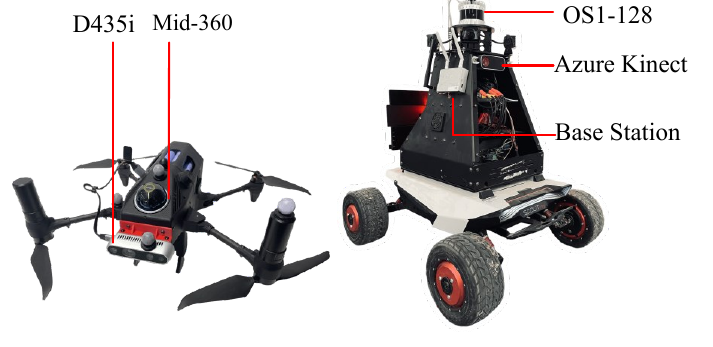}
\caption{Hardware Configurations of UAV and UGV.}
\label{fig:platform}
\end{figure}

The experimental platform consists of two main components: UGV and UAV (Fig.\ref{fig:platform}). The UGV utilizes a robust AgileX Scout Mini, equipped with an Azure Kinect camera, an Ouster LiDAR OS-128 for comprehensive environmental perception, and an onboard mini PC featuring an Intel i9-13900H processor alongside an NVIDIA RTX 4070 GPU with 8GB of VRAM. The UAV is outfitted with a Pixhawk6 Pro flight control module for enhanced stability and precision. It integrates a RealSense D435i for visual sensing and a Livox Mid-360 LiDAR to facilitate comparative analysis with LiDAR-based relative positioning methods. The core computing unit of the UAV is a Jetson Orin NX, boasting 16GB of VRAM. In addition, the communication for our air-ground collaboration utilizes a video transmission device with a bandwidth of 40 Mbps. This setup enables rigorous testing and validation of our methodologies in diverse scenarios. 
\subsection{Dataset}
We provide two datasets corresponding to scenarios with varying levels of co-visibility: sufficient co-visibility 
 \textit{Dataset1} and no co-visibility \textit{Dataset2} at the initial state. Each dataset includes synchronized camera, LiDAR, and IMU data, along with pre-calibrated parameters. The calibration data comprises camera intrinsics, as well as extrinsic parameters between the camera, LiDAR, and IMU. To ensure high-quality ground truth, we employ the FZMotion Motion
Capture System to record the pose trajectories of both the UAV and UGV during their operations. The presence of comprehensive sensor data and accurate calibration parameters facilitates benchmarking across different approaches, such as single-modal-based and cross-modal-based methods. The inclusion of ground truth trajectories enables precise assessment of reconstruction accuracy and relative pose estimation, making the datasets valuable for advancing research in air-ground collaboration, especially in GPS-denied scenarios.
\subsection{Metrics}
We assume that the gravity estimation at both the UAV and UGV ends is accurate, with the z-axis of each map aligned parallel to the gravity direction. Consequently, our relative pose estimation consists of only 4 DoF: x, y, z, and yaw. During quantitative evaluation, translational error is measured using the Euclidean distance (m) in three-dimensional space, while rotational error is assessed solely by the deviation of the yaw angle ($^{\circ}$). The duration of state estimation (ms) serves as a metric for evaluating algorithm efficiency. In comparison with other methods, the stability and robustness of the algorithm are evaluated based on the success rate (SR) of pose estimation. In tables presented in the $\mathcal{A}/\mathcal{B}$ format, $\mathcal{A}$ denotes the number of successful trials, while $\mathcal{B}$ represents the total number of tests conducted.

\subsection{Ablation Study}
The cost function $\mathbf{r}_{W_GW_A}$ avoids the instability of state estimation dependent on a single data pair, though it offers limited improvement in accuracy. $\mathbf{r}_{C_GC_A}$ leverages the spatial consistency in the common-view region between the UGV and UAV for local BA to refine the relative pose. $\mathbf{r}_{G_C}$ takes into account errors caused by the extrinsic parameters, time offset, and pixel-cloud correspondence between sensors on the UGV side, and corrects these errors through local BA. The results in Table \ref{tab:ab} further substantiate our claims, highlighting the rationality and superiority of our method across different datasets. Nevertheless, as the complexity of the optimization process increases, while the accuracy can reach centimeter or sub-centimeter levels, the computational time also increases by more than 100 times. In practical experiments, however, performing relative localization once per second is fully sufficient to meet the requirements, achieving a favorable balance between precision and efficiency.
 
\begin{table}[t!]
\renewcommand\arraystretch{1.2}
\setlength{\tabcolsep}{3pt}
\centering
\caption{evaluation for state estimation of different settings}
\label{tab:ab}
\begin{tabular}{l|ccc|ccc}
\hline
  \begin{tabular}[c]{@{}c@{}}Dataset\end{tabular} &
  \begin{tabular}[c]{@{}c@{}}r$_{G_C}$\end{tabular} &
  \begin{tabular}[c]{@{}c@{}}r$_{C_GC_A}$\end{tabular} &
  \begin{tabular}[c]{@{}c@{}}r$_{W_GW_A}$\end{tabular} &
  \begin{tabular}[c]{@{}c@{}}Translational\\ Error (\textit{m})\end{tabular} &
  \begin{tabular}[c]{@{}c@{}}Rotation\\ Error (\textit{$^\circ$})\end{tabular} &
  \begin{tabular}[c]{@{}c@{}}Time (\textit{ms})\end{tabular}
  \\ \hline
 \multirow{3}{*}{\textit{Dataset1}}&$\bm{\times}$&$\bm{\times}$&\checkmark&0.187&6.42&\textbf{3.86} \\
 &$\bm{\times}$&\checkmark&\checkmark&0.137&6.04&311 \\
&\checkmark&\checkmark&\checkmark&\textbf{0.065}&\textbf{5.36}&462 \\
 \hline
 \multirow{3}{*}{\textit{Dataset2}}&$\bm{\times}$&$\bm{\times}$&\checkmark&0.129&6.24&\textbf{5.27} \\
 &$\bm{\times}$&\checkmark&\checkmark&0.090&6.20&307 \\
&\checkmark&\checkmark&\checkmark&\textbf{0.004}&\textbf{5.07}&556 \\
\hline
\end{tabular}
\end{table}

\subsection{Performance with Different Keypoint Counts}
Limited communication bandwidth in air-ground collaborative systems can significantly impact the performance of relative localization. In our system, the UAV transmits per-frame data packets to the UGV containing four principal components: $n$ 2D keypoint coordinates and 64D feature descriptors, 7-DoF pose representation (quaternion + translation), 512D global descriptor. Given 4-byte quantization per dimension, the total bandwidth demand per second is formulated as:
\begin{equation}
    \mathcal{W} = 4 \times (2n + 64n + 7 + 512) \text{ bytes} \quad \Rightarrow \quad \frac{\mathcal{W} \times 8}{10^6} \text{ Mbps}
\end{equation}
As shown in the Table \ref{tab:keypoint}, the increase in the number of keypoints leads to a more significant improvement in accuracy, but this also results in higher computation time. A total of 1,024 keypoints is sufficient to achieve sub-centimeter-level accuracy, while requiring a bandwidth of only 2.18 Mbps. In scenarios where lower precision is acceptable, our bandwidth requirement can be further reduced to less than 0.3 Mbps. This demonstrates an efficient trade-off between precision and bandwidth consumption, highlighting the effectiveness of our approach in resource-constrained environments.
\begin{table*}[htbp!]
\centering
\caption{Performance with different keypoint counts}
\begin{tabular}{@{}cccccccc@{}}
\toprule
\multirow{2}{*}{\makecell{Keypoint \\Counts}} & \multicolumn{3}{c}{\textit{Dataset1}} & \multicolumn{3}{c}{\textit{Dataset2}} &\multirow{2}{*}{\makecell{Bandwidth\\(Mbps)}} \\ \cmidrule(lr){2-4} \cmidrule(l){5-7}& Rotation Error($^\circ$) & {Trans Error(m)} & {Time(ms)} & {Rotation Error($^\circ$)} & {Trans error(m)} & {Time(ms)} \\ \midrule
128& 6.60&0.341 &115&8.56&0.204&73&0.29\\ 
256& 6.29  & 0.305& 253& 10.97 &0.112&182&0.56\\ 
512& 9.68& 0.209& 346& 8.76& 0.052& 445&1.10\\
1024& 5.36 & 0.065& 462 & 5.07& 0.004 &556&2.18\\ \bottomrule
\end{tabular}
\label{tab:keypoint}
\end{table*}
\subsection{Evaluation for Database Maintenance and Retrieval}
As an alternative to DBoW3, our method achieves better compatibility with neural network-based global descriptors. We evaluate both methods across three key metrics: average single-frame incremental insertion time (SIT), including feature extraction and database construction; average retrieval time per single frame (SRT); and accuracy (Acc). Using our actual recorded data, we incrementally constructed a database and retrieved the most similar candidate, as illustrated in Table \ref{tab:bt}.
The SIT differs by approximately 1 ms between the two methods, with both ensuring real-time performance. Specifically, our method has an SIT of 26.36 ms compared to DBoW3's 25.16 ms. In terms of accuracy, our method exhibits a slight advantage, achieving about a 2\% improvement over DBoW3, with accuracies of 82\% and 80\%, respectively. Regarding retrieval efficiency, our method demonstrates a substantially greater advantage, completing retrieval within 0.84 ms—approximately 25 times faster than DBoW3, which completes retrieval in 21.14 ms. These results highlight the superior efficiency and effectiveness of our proposed approach.
\begin{table}[t!]
\centering
\caption{DboW3 VS Ours}
\begin{tabular}{@{}lccc@{}}
\toprule
Methods&SIT (ms)&SRT (ms)&Acc (\%) \\ \midrule
Ours&26.36&\textbf{0.84}&\textbf{82}\\ 
DBoW3&\textbf{25.16}&21.14& 80 \\ 
\bottomrule
\end{tabular}
\label{tab:bt}
\end{table}
\subsection{Comparasion with Other Methods}

We compared our approach against state-of-the-art baselines. We selected D$^2$SLAM \cite{xu2024d}, a leading vision-based multi-robot SLAM, as the primary benchmark.  We also deployed SlideSLAM \cite{slideslam}, a state-of-the-art multi-robot LiDAR SLAM method that heavily relies on semantic information, but it failed to achieve relative localization. Consequently, we chose TEASER \cite{teaser}, one of the SOTA point cloud registration algorithm integrated within hdl\_localization \cite{koide2019portable}, for evaluation. As shown in Table \ref{tab:baseline}, D$^2$SLAM relies heavily on sufficient common-view information during initialization. Therefore, it cannot perform relative localization in \textit{Dataset2} without common-view data in the initial state. Even with sufficient common-view data, its performance remains unstable, with limited accuracy. While TEASER achieves favorable results, it requires the pre-computation of FPFH descriptors. Moreover, this approach has specific requirements for point cloud density parameters and map scale, and failure to satisfy these conditions may lead to localization failure. Our method achieves the best balance in terms of stability, accuracy, and efficiency.
\begin{table}[t!]
\renewcommand\arraystretch{1.2}
\setlength{\tabcolsep}{3pt}
\centering
\caption{evaluation for state estimation with different methods}
\label{tab:baseline}
\begin{tabular}{l|lcccc}
\hline
\begin{tabular}[c]{@{}c@{}}Dataset\end{tabular} &
  \begin{tabular}[c]{@{}c@{}}Methods\end{tabular} &
  \begin{tabular}[c]{@{}c@{}}Translational\\ Error (\textit{m})\end{tabular} &
  \begin{tabular}[c]{@{}c@{}}Rotation\\ Error (\textit{$^\circ$})\end{tabular} &
   \begin{tabular}[c]{@{}c@{}}SR\end{tabular} &
  \begin{tabular}[c]{@{}c@{}}Time (\textit{ms})\end{tabular}
  \\ \hline
 \multirow{3}{*}{\textit{Dataset1}} & D$^2$SLAM&0.554&66.37&2/10&\textbf{27.2} \\
 &FPFH-Teaser&\textbf{0.039}&7.162&6/10&5579 \\
 &Ours&0.065&\textbf{5.361}&\textbf{10/10}&465 \\
 \hline
  \multirow{3}{*}{\textit{Dataset2}} & D$^2$SLAM&$\bm{\times}$&$\bm{\times}$&$\bm{\times}$&$\bm{\times}$ \\
 &FPFH-Teaser&0.029&5.693&7/10&4963 \\
 &Ours&\textbf{0.004}&\textbf{5.066}&\textbf{10/10}&\textbf{556}\\
 \hline
\end{tabular}
\end{table}



\section{CONCLUSIONS}
In conclusion, this paper presents a novel semi-distributed cross-modal air-ground relative localization framework that decouples from the state estimation of multiple robots, enhancing both flexibility and accuracy in collaborative tasks. By leveraging the UGV's ability to integrate diverse sensors, the proposed method independently performs SLAM on both the UGV and UAV while utilizing deep learning-based keypoint descriptors and global descriptors for localization. The two-stage optimization process ensures rapid and accurate relative pose estimation, while the database maintenance method ensures incremental and real-time keyframe insertion and retrieval. Experimental results demonstrate significant improvements in localization performance, while the keypoint-based communication strategy effectively reduces bandwidth usage, making the approach suitable for bandwidth-constrained environments. This work provides a promising direction for future air-ground collaboration in GPS-denied and bandwidth-limited scenarios, with potential applications in both wilderness and post-disaster settings.

\bibliographystyle{plain}  
\bibliography{ref} 
\end{document}